\documentclass[letterpaper, 10 pt, journal, twoside]{ieeetran}
\usepackage{makecell} 
\usepackage{multirow}

\usepackage{cite}
\usepackage{amsmath,amssymb,amsfonts}
\usepackage[ruled,vlined,linesnumbered]{algorithm2e}
\usepackage{graphicx}
\usepackage{textcomp}
\usepackage{xcolor}
\usepackage{times}
\usepackage{colortbl} 

\usepackage{pifont}
\usepackage{array}
\usepackage{booktabs} 
\usepackage{siunitx}
\usepackage{tabularx}
\usepackage{subcaption}
\usepackage{orcidlink}
\usepackage{enumitem, amsthm}

\newcommand{\A}{\mathcal{A}}
\newcommand{\cT}{\mathcal{T}}
\newcommand{\cE}{\mathcal{E}}
\newcommand{\p}{\mathbf{p}}
\newcommand{\x}{\mathbf{x}}

\newcommand{\cR}{\mathcal{R}}

\newcommand{\cF}{\mathcal{F}}
\newcommand{\cG}{\mathcal{G}}
\newcommand{\cN}{\mathcal{N}}
\newcommand{\Rn}{\mathbb{R}}

\providecommand{\abs}[1]{\lvert#1\rvert}
\renewcommand{\b}[1]{\ensuremath{\mathbf{#1}}} 
\newcommand{\norm}[1]{\ensuremath{\left\|#1\right\|}} 

\def \x {{\b{x}}}
\def \k {{\b{k}}}

\def \Rn {{\mathbb{R}}}

\theoremstyle{remark}
\newtheorem{rem}{\bf Remark}
\newtheorem{theorem}{Theorem}

\begin{document}

\title{Communication and Energy-Aware Multi-UAV Coverage Path Planning for Networked Operations}

 \author{Mohamed~Samshad\orcidlink{0009-0007-0298-4831},
 and~Ketan~Rajawat\orcidlink{0000-0002-4508-0062},~\IEEEmembership{Member,~IEEE}
\thanks{


The authors are with the Department of Electrical Engineering, Indian Institute of Technology Kanpur, Kanpur 208016, India (e-mail: msamshad@iitk.ac.in; ketan@iitk.ac.in).

}
\thanks{Manuscript received [XYZ]; date of current version April 20, 2025. }}

\markboth{IEEE ROBOTICS AND AUTOMATION LETTERS, PREPRINT VERSION}%
{IEEE ROBOTICS AND AUTOMATION LETTERS, PREPRINT VERSION}



\maketitle

\begin{abstract}
This paper presents a communication and energy-aware multi-UAV Coverage Path Planning (mCPP) method for scenarios requiring continuous inter-UAV communication, such as cooperative search and rescue and surveillance missions. Unlike existing mCPP solutions that focus on energy, time, or coverage efficiency, the proposed method generates coverage paths that minimize a specified combination of energy and inter-UAV connectivity radius. Key features of the proposed algorithm include a simplified and validated energy consumption model, an efficient connectivity radius estimator, and an optimization framework that enables us to search for the optimal paths over irregular and obstacle-rich regions. The effectiveness and utility of the proposed algorithm is validated through simulations on various test regions with and without no-fly-zones. Real-world experiments on a three-UAV system demonstrate the remarkably high 99\% match between the estimated and actual communication range requirement. 
\end{abstract}


\begin{center}
\vspace{-0.2cm}
SUPPLEMENTARY MATERIAL
\end{center}
\vspace{-0.2cm}
\noindent Video: {\footnotesize \url{https://youtu.be/9acU-A5RCKg}}\\
\noindent Code: {\footnotesize \url{https://gitlab.com/samshadnc/communication-aware-mcpp}}

\section{Introduction}
Coordinated missions performed by teams of Unmanned Aerial Vehicles (UAVs) are becoming increasingly critical for addressing the complex operational challenges encountered in a wide range of scenarios. Multi-UAV systems enable quick completion of complicated missions, such as in surveillance \cite{survelliance}, photogrammetry \cite{photog}, agricultural monitoring, precision farming \cite{agri1, crophealth, cropferti}, and Search And Rescue (SAR) \cite{sar, sar1}. Compared to a single-UAV operation, a multi-UAV system can enhance the mission efficiency and reliability by providing increased coverage area, reduced operational time, improved redundancy, and enhanced collaboration capabilities. 

Inter-UAV communication is the cornerstone of many such coordinated missions, e.g. in SAR operations where the UAVs may need to share the locations of victims or environmental hazards, or in surveillance operations requiring cooperative target tracking. For multi-UAV systems that are connected over a mesh network, maintaining inter-UAV communication requires ensuring that the UAVs do not stray too far from each other. In practice, UAVs are generally equipped with small antennas, and therefore the inter-UAV communication range is typically shorter  than the UAV-ground control station range. Hence, planning multi-UAV paths without considering inter-UAV communication requirements may result in a fragmented UAV network that is unable to complete the mission. 

We consider the Multi-UAV Coverage Path Planning (mCPP) problem, where the goal is to generate multi-UAV paths \emph{covering} a specified area. Such coverage requirement arises in a number of UAV applications, e.g., photogrammetry, surveillance, environmental monitoring, sensing, and precision farming. The mCPP problem has been widely studied and enhanced algorithms have been proposed to handle No-Fly Zones (NFZs) \cite{kapoutsis2017darp, energy_aware, choset1998coverage}, enhance energy efficiency \cite{energy_aware, stefanopoulou2024improving}, optimize flight speed \cite{energy_aware}, and support uneven workload distribution \cite{kapoutsis2017darp, stefanopoulou2024improving}. However, existing works in the literature do not consider the communication requirements. 

In this paper, we consider the communication and energy-aware mCPP problem which seeks to find the coverage path that minimizes a combination of energy consumption and connectivity radius. The framework relies on two key building blocks: a light weight energy consumption model and an connectivity radius estimator. The estimates allow us to evaluate candidate paths and search for the optimal path. 


To construct candidate paths that support irregular and obstacle-rich regions, we utilize the Divide Areas based on Robots initial Positions (DARP) algorithm \cite{kapoutsis2017darp} for dividing the coverage workload among the given UAVs and the Spanning Tree Coverage (STC) algorithm \cite{gabriely2001spanning} for generating the paths for each UAV. Subsequently, we utilize Bayesian optimization to find the optimal path that minimizes the specified combination of energy and connectivity radius. In summary, the key contributions of this work are as follows: 
\begin{itemize}
    \item We propose a parameterizable constant-regime energy model that can be analytically evaluated for any path, achieving less than 5\% error in practice.  
    \item We prove that the mesh-network connectivity radius is Lipschitz continuous, enabling use of the Shubert–Piyavskii global optimizer and yielding significant speedups, especially for larger UAV fleets.  
    \item We develop a nested Bayesian optimization strategy that efficiently finds near-global optima by pruning candidates using a running median heuristic.  
    \item The resulting algorithm is a complete, end-to-end planner that supports irregular and obstacle-rich regions of interest (ROIs) with guaranteed coverage.  
\end{itemize}

\section{RELATED WORK}

\begin{figure*}[!ht]
    \centering
    \includegraphics[width=\textwidth]{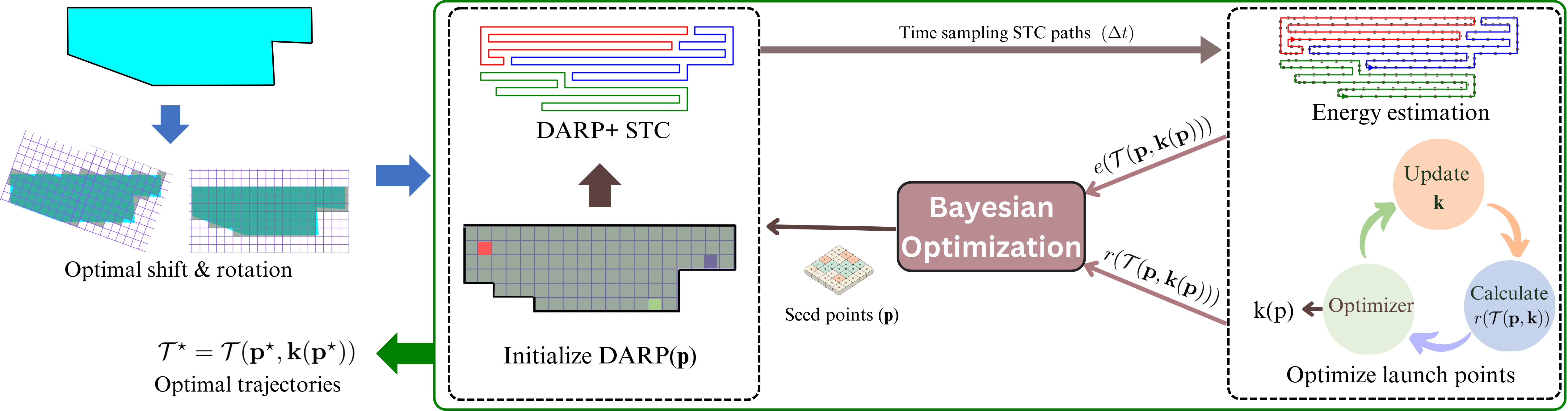}
    \caption{Optimization process breakdown: ROI Discretization through optimal shifts and rotation, DARP-MST path generation, time-sampling \& energy estimation, launch point optimization by connectivity analysis, Bayesian optimization with range and energy for next DARP initialization}
    \label{fig:flow}
\end{figure*}

Coverage Path Planning (CPP) is a well-studied problem, with several solutions already implemented in real-world path planning applications such as floor-cleaning robots, painting systems, UAV survey path planners, and agricultural spraying robots\cite{ardupilot_mission_planner, irobot_website, dji_flight_planner}. This section discusses some of the works that have provided key motivation for this paper through their innovative approaches to the problem or optimized results for different flight parameters.

A number of CPP techniques are reviewed in \cite{E_Galceran(2013), cabreira2019survey}, including bio-inspired, computational, and analytical approaches for both simple and complex ROIs. In this context, it is observed that a number of techniques utilize variants of trapezoidal cell decomposition \cite{cabreira2019survey}, STC \cite{kapoutsis2017darp}, and energy-aware CPP strategies \cite{energy_aware}, with varying emphasis on optimizing flight parameters such as energy consumption, path length, and coverage efficiency. 

Boustrophedon cell decomposition is an alternative to classical trapezoidal decomposition \cite{choset1998coverage, trapezoidal} that enables CPP in ROIs with no-fly zones (NFZs) by partitioning them into cells and using systematic back-and-forth sweeping paths within each cell. A graph traversal method is then used to connect these individual paths into a complete coverage path. Building on this, \cite{bahnemann2021revisiting} revisits boustrophedon CPP by replacing the graph traversal method with the Generalized Traveling Salesman Problem (GTSP) \cite{gtsp} to determine the optimal path configuration, minimizing time and directional changes and improving energy efficiency.

While these studies present foundational CPP algorithms, many proposed solutions are optimized for various flight parameters, focusing on minimizing flight path length \cite{energy_aware,apostolidis2022cooperative, popcorn}, mission time \cite{bahnemann2021revisiting}, the number of path turns \cite{li2011coverage, vandermeulen2019turn, stefanopoulou2024improving}, energy consumption along the path \cite{energy_aware, stefanopoulou2024improving}, and inter-UAV connectivity \cite{formation}. A recent study introduces an Energy-Aware mCPP (EA-mCPP) strategy specifically for multi-UAV operations, focusing on optimizing energy consumption and flight speed \cite{energy_aware}. This approach uses an accurate energy model \cite{energy_cal} to predict expenditure across boustrophedon cells and calculate optimal flight speeds. The method employs the Multiple Set Traveling Salesman Problem (MS-TSP) \cite{mstsp} to determine energy-efficient routing, ensuring effective resource allocation and path utilization in UAV missions.
\begin{table}[t]
\centering
\scriptsize
\renewcommand{\arraystretch}{1.4} 
\setlength{\tabcolsep}{3pt} 
\caption{Qualitative Comparison of the Related Methods}
\vspace{0.2cm} 

\arrayrulecolor{black} 
\begin{tabular}{p{3cm} >{\centering\arraybackslash}p{1.1cm} >{\centering\arraybackslash}p{1.10cm} >{\centering\arraybackslash}p{1.10cm} >{\centering\arraybackslash}p{1cm}}
\hline
 & Comm. aware & Energy aware & Custom load dist. & No-fly zones \\ 
\hline
Proposed & \ding{51} & \ding{51} & \ding{51} & \ding{51} \\ 
\arrayrulecolor[gray]{0.7}\hline 
\begin{tabular}[c]{@{}l@{}}Cooperative mCPP\cite{apostolidis2022cooperative}\end{tabular} & \ding{55} & \ding{51} & \ding{51} & \ding{51} \\ 
\hline
\begin{tabular}[c]{@{}l@{}}EA-mCPP\cite{energy_aware}\end{tabular} & \ding{55} & \ding{51} & \ding{55} & \ding{51} \\ 
\hline
\begin{tabular}[c]{@{}l@{}}DARP+MST\cite{kapoutsis2017darp}\end{tabular} & \ding{55} & \ding{55} & \ding{51} & \ding{51} \\ 
\hline
\begin{tabular}[c]{@{}l@{}}O-DARP\cite{stefanopoulou2024improving}\end{tabular} & \ding{55} & \ding{51} & \ding{51} & \ding{51} \\ 
\hline
\begin{tabular}[c]{@{}l@{}}POPCORN+SALT\cite{shah2022large}\end{tabular} & \ding{55} & \ding{51} & \ding{51} & \ding{55} \\ 
\hline
\begin{tabular}[c]{@{}l@{}}Formation flying\cite{formation}\end{tabular} & \ding{51} & \ding{51} & \ding{55} & \ding{55} \\
\arrayrulecolor{black}\hline 
\end{tabular}
\arrayrulecolor{black} 
\label{qualilty_comp}
\end{table}
\setlength{\textfloatsep}{4pt}

All of the discussed CPP solutions utilize area partitioning and optimal allocation. In \cite{formation}, a formation flying technique was proposed for multi-UAV coverage missions in ROIs without any NFZs. This approach ensures that the inter-UAV distance remains within the range due to the rigid formation shape, thereby maintaining the UAVs within communication range of each other. Another distinct technique involves discretizing the ROI into grid cells and performing STC. In \cite{kapoutsis2017darp}, a fast and effective algorithm is proposed by discretizing the ROI into square grid cells and then using a modified Voronoi algorithm to allocate cells to each robot. The STC algorithm \cite{gabriely2001spanning} is then implemented for each region to generate independent coverage paths. A similar approach is used in \cite{zhang2023multi}, where adaptive grid size discretization and the optimal allocation Minimum Spanning Tree (MST) are employed to generate smooth and minimal-cost paths. More recently, machine learning techniques have also been used to generate energy-efficient coverage paths \cite{stefanopoulou2024improving}. In this approach, the energy consumption of paths is mapped to the number of turns, and optimization techniques are applied to generate paths with fewer turns across different DARP initializations. We remark that many of these techniques cannot handle complex ROIs and none of them consider communication or range requirements. Finally, methods that involve formation flying \cite{formation}, focus solely on maintaining the inter-UAV distance and cannot work for ROIs with NFZs. A comparison of the discussed approaches is summarized in Table \ref{qualilty_comp}.
\section{PROBLEM FORMULATION}\label{problem_formulation}
We consider the problem of surveying a two-dimensional ROI $\A \subset \Rn^2$. The mission involves deploying a fleet of $N$ multirotor UAVs, each operating within the 2D plane above the ground. The position of the $i$-th UAV at time $t$ is denoted by $\x_i(t) \in \mathbb{R}^2$ and the set of positions of all UAVs over a horizon of $T$ seconds is referred to as a multi-UAV trajectory $\cT:=\{\x_i(t) \mid 1\leq i \leq N, t \in [0,T]\}$. Each UAV is equipped with a sensor capable of observing an area on the ground, referred to as the sensor footprint and denoted by $\cF(\x_i(t)) \subset \Rn^2$. The coverage requirement stipulates that every point within the ROI must be observed at least once by one of the UAV's sensor during the mission, so that
\begin{align}\label{cov}
 \bigcup_{i=1}^N \bigcup_{t\in[0,T]} \cF(\x_i(t)) = \A. 
\end{align}
Notably, we do not require $\A$ to be convex or simply connected, and NFZs may exist within the convex hull of $\A$. The total energy consumed by all the $N$ UAVs when following a given multi-UAV trajectory $\cT$ is denoted by $e(\cT)$. The UAVs are equipped with omnidirectional antennas and form a mesh network to maintain continuous connectivity \cite{draves2004routing}. Given $\cT$, the communication radius required to ensure that all the UAVs can communicate with each other, either directly or through multi-hop paths involving other UAVs, is referred to as the \emph{connectivity radius} and denoted by $r(\cT)$. 

Solving the coverage problem will broadly require solving two sub-problems. Firstly, we want to generate a large number of collision-free trajectories  that satisfy the coverage condition in \eqref{cov}. Secondly, among the trajectories that satisfy \eqref{cov}, we need to chose the trajectory $\cT$ that minimizes $f_o(\cT) := r(\cT) + \lambda e(\cT)$, where $\lambda$ is a weighting parameter that balances the relative importance of the required communication range and total energy consumption. In practice, however, we integrate these two phases, as the proposed algorithm will simultaneously generate and prune trajectories based on their objective function values.

Before describing the algorithm, we note that both $r(\cT)$ and $e(\cT)$ must be estimated for candidate trajectory $\cT$ and the next two subsections describe the estimation routines used. While the resulting estimates are generally approximate---since they are oblivious to the environment---they are crucial because they enable us to systematically explore the space of valid trajectories. 

\subsection{Estimation of total energy \texorpdfstring{$e(\cT)$}{e(T)}}\label{enest}

There exist a number of approaches for estimating the energy consumption of UAV trajectories. For instance, the approach in \cite{energy_cal} provides accurate energy estimates by utilizing detailed models for the UAV aerodynamics, motor, and battery. Although these detailed models can deliver precise energy estimates if available, they necessitate estimating a large number of UAV-specific parameters. Additionally, their high computational complexity makes it challenging to integrate them into the current path planning and optimization framework. Since UAVs maintain a constant speed for the majority of their flight path, we put forth a lightweight energy estimation algorithm that turns out to be sufficiently accurate. 

The proposed energy estimation algorithm operates under two key modeling simplifications (a) the UAV paths consist solely of hovering, straight line segments covered at constant speed, and sharp turns, and (b) the UAV consumes constant powers $P_h$, $P_f$, and $P_t$ during hovering, forward flight, and turns, respectively. As we shall see later, the proposed algorithm will generate paths that adhere to the first assumption. The second simplification also introduces errors, e.g., due to wind, but the cumulative error is assumed to be negligible and ignored for the sake of simplicity. We further note that the more complicated energy estimation methods in \cite{energy_cal, energy_aware} also suffer from errors due to the presence of unknown winds. 

Under these simplifications and for a given trajectory $\cT_i$ of the $i$-th UAV, we simply need to measure the total time duration for which the UAV hovers ($T_h(\cT_i)$), moves along a straight line path ($T_f(\cT_i)$), and makes a sharp turn ($T_t(\cT_i)$) such that $T_h(\cT_i)+T_f(\cT_i)+T_t(\cT_i) = T$. The total energy consumption for trajectory $\cT$ is given by 
\begin{align}
    e(\cT) = \sum_{i=1}^N P_hT_h(\cT_i) + P_fT_f(\cT_i) + P_tT_t(\cT_i).
\end{align}

The energy estimation model was validated through systematic real-world experimentation. Since the mission trajectories consist of only straight-line segments and turning maneuvers, we excluded the take-off and landing phases from our analysis. During validation, we recorded the UAV's power consumption throughout the mission duration, enabling characterization of distinct power profiles for straight-line flight ($P_f$) and turning maneuvers ($P_t$). The total energy was then computed by integrating these profiles over their respective time intervals. While this simplified approach proves effective, its accuracy critically depends on two factors: the precision of power consumption calibration and the adherence of actual missions to our modeling assumptions. Notably, the model assumes constant-speed straight-line flight, symmetric turning maneuvers, and environmental conditions that closely match the calibration phase parameters. The model may require recalibration or exhibit reduced accuracy in scenarios involving variable flight speeds, asymmetric maneuvers, or significant environmental disturbances that deviate from the calibration conditions. Despite these constraints, the validation framework achieves a practical balance between computational efficiency and accuracy for our specific mission profiles.
To verify the estimation model, we conducted three experiments on simple polygonal ROIs using a 3-UAV system. Test results showed total energy consumption of 134.44 Wh, 137.04 Wh, and 136.82 Wh, with individual UAV consumption ranging from 42.12-46.82 Wh per mission. Comparing with our estimated 130.44 Wh yielded an average accuracy of 95.8\% across all tests, validating the model's reliability for multi-UAV mission planning.

\subsection{Estimation of connectivity radius \texorpdfstring{$r(\cT)$}{r(T)}}\label{radest}
\label{subsec: connectivity_radius}
We begin with formally defining $r(\cT)$ for a network of UAVs, represented by a time-varying weighted complete graph $\mathcal{G}(t) = (\mathcal{N}, \mathcal{E}, w_t)$. The set of nodes $\mathcal{N} = \{1, 2, \ldots, N\}$ represents the $N$ UAVs, and the edge set $\mathcal{E} = \{(i, j) \mid i, j \in \mathcal{N},\, i \neq j\}$ includes all possible pairs of UAVs, reflecting potential communication links. The graph is dynamic due to its time-dependent weight function $w_t: \mathcal{E} \rightarrow \mathbb{R}_+$; specifically, the weight of edge $(i, j)$ at time $t$ is given by $w_t(i, j) = \| \x_i(t) - \x_j(t) \|$. 
 
Overloading the notation, let $r(t)$ represent the minimum communication radius required to ensure that the mesh network formed by the UAVs is connected at time $t$. Equivalently, 
\begin{align}
r(t) = \min \left\{ r \geq 0 \mid \text{the subgraph } \mathcal{G}_r(t) \text{ is connected} \right\}.
\end{align}
where $\cG_r(t):=(\cN, \cE_r(t))$ is the subgraph induced by including only the edges with weights less than or equal to $r$, so that 
\begin{align}
\mathcal{E}_r(t) = \left\{ (i, j) \in \mathcal{E} \mid w_t(i, j) \leq r \right\}.
\end{align}
In other words, $\mathcal{E}_r(t)$ consists of all edges that connect UAVs that are within $r$ of each other at time $t$. The subgraph $\mathcal{G}_r(t)$ is thus the graph where UAVs can communicate if they are within range $r$. Finally, the connectivity radius for the entire trajectory $\mathcal{T}$ is given by $r(\cT) = \max_{t \in [0, T]} r(t)$.

Given the graph $\cG(t)$, we can determine $r(t)$ by identifying the minimum bottleneck spanning tree (MBST) of $\cG(t)$. An MBST  is a spanning tree in which the largest edge weight, known as the 'bottleneck' edge, is minimized across all possible spanning trees of the graph. In general, for a complete graph, the MBST can be found using Camerini's algorithm, which operates in $O(N^2)$ time \cite{camerini1978min}. This approach is computationally feasible when $N$ is relatively small, typically manageable for $N$ below 10. Camerini's algorithm identifies the MBST by iteratively selecting edges to form a spanning tree while minimizing the maximum edge weight, which aligns with the concept of reducing the "bottleneck."

However, when $N$ is large, an alternative method is required to maintain computational efficiency. In particular, $\mathcal{G}(t)$ is a Euclidean graph, as the weights $w_t(i, j)$ are derived from Euclidean distances. For Euclidean graphs, a minimum spanning tree (MST) can be efficiently computed in $O(N \log(N))$ time using Delaunay triangulation. Once the MST of $G(t)$ is computed, $r(t)$ can be found directly from the weight of the heaviest edge in the MST, which, by definition, is also the MBST.

To determine $r(\cT)$, we need to solve the global scalar optimization problem $\max_{t \in [0,T]} r(t)$. The grid-search algorithm, which finds $\max_{1\leq k \leq T/{\Delta t}} r(k\Delta t)$, is efficient for small $T$. However, for large $T$, we require a more efficient search algorithm that takes advantage of the special structure of the problem. To this end, we have the following result: 
\begin{theorem}\label{thlip}
The communication radius $r(t)$ is $2v$-Lipschitz continuous, i.e., $|r(t) - r(t')| \leq 2v |t - t'|$ for any $t, t' \in [0,T]$.
\end{theorem}
\begin{IEEEproof}
We begin with introducing some notation. Let $\cG_\ell(t) = (\cN, \cE_\ell(t))$ denote the $\ell$-th spanning tree of $\cG(t)$, where $\cN$ is the set of UAVs (which remains constant over time), and $\cE_\ell(t)$ is the set of edges in that spanning tree at time $t$. Let $r_\ell(t) := \max_{(i,j) \in \cE_\ell(t)} \norm{\x_i(t)-\x_j(t)}$ denote the weight of its bottleneck (longest) edge. By definition, $
r(t) = \min_{\ell} r_\ell(t)$. Let $\ell^\star(t) = \arg\min_{\ell} r_\ell(t)$ denote the index of the MBST at time $t$, so $ r(t) = r_{\ell^\star(t)}(t)$.

The proof begins with bound $r(t)-r(t')$, which can be written as:
\begin{align}
r(t) - r(t') &= r_{\ell^\star(t)}(t) - r_{\ell^\star(t')}(t') \\
&\hspace{-10mm}= [ r_{\ell^\star(t)}(t) - r_{\ell^\star(t')}(t) ] + [ r_{\ell^\star(t')}(t) - r_{\ell^\star(t')}(t') ] \\
&\leq [ r_{\ell^\star(t')}(t) - r_{\ell^\star(t')}(t') ],
\end{align}
where the inequality follows because $ r_{\ell^\star(t)}(t) \leq r_{\ell^\star(t')}(t) $, as $ \ell^\star(t) $ is the MBST at time $ t $ and therefore has the smallest bottleneck edge weight among all spanning trees at that time. Note that $\ell^\star(t')$ is also a spanning tree but not necessarily the MBST at time $t$.

Next, consider the term $ r_{\ell^\star(t')}(t) - r_{\ell^\star(t')}(t') $. Since $ \cE_{\ell^\star(t')} $ is the edge set of a specific spanning tree (the MBST at time $ t' $), and the set of UAVs $\cN$ remains the same, the edge set $ \cE_{\ell^\star(t')} $ does not change between times $t'$ and $t$; only the positions of the nodes (and thus the edge weights) change. Therefore:
\begin{align}
&r_{\ell^\star(t')}(t) - r_{\ell^\star(t')}(t') \nonumber\\
&= \max_{(i,j) \in \cE_{\ell^\star(t')}} \norm{\x_i(t) - \x_j(t)} - \max_{(i,j) \in \cE_{\ell^\star(t')}} \norm{\x_i(t') - \x_j(t')} \nonumber\\
&\leq \max_{(i,j) \in \mathcal{E}_{\ell^\star(t')}} \left| \| \mathbf{x}_i(t) - \mathbf{x}_j(t) \| - \| \mathbf{x}_i(t') - \mathbf{x}_j(t') \| \right|,
\end{align}
where we used the fact that the difference between the maxima of two sets is less than or equal to the maximum absolute difference between corresponding elements, i.e., $\max_k a_k - \max_k b_k \leq \max_k | a_k - b_k |$. Note also that since the edge sets are the same at times $t$ and $t'$, we can directly compare the edge weights.

It remains to bound the change in edge weights of a particular edge. Since each UAV moves at most $v \abs{ t - t'}$ units over the time interval $\abs{t - t'}$, the change in the length of any edge $(i, j) $ is upper bounded by $2v\abs{t - t'}$. Hence we have been able to establish that $r(t)-r(t')\leq 2v\abs{t-t'}$. Finally, by interchanging $ t $ and $ t' $ in this argument, we obtain $r(t') - r(t) \leq 2v\abs{t - t'}$. Combining both inequalities, we obtain the desired Lipschitz continuity result. 
\end{IEEEproof}

Having established the Lipschitz continuity of $r(t)$, it is now possible to use the Shubert–Piyavskii method for finding its minimum. The Shubert–Piyavskii method is a one-dimensional minimization algorithm that can find the global minimum of a $2v$-Lipschitz continuous function up to an accuracy of $\epsilon$ within at most $\mathcal{O}(vT\epsilon^{-1})$ function evaluations, where the computational complexity of each iteration is fixed. Note that the grid search algorithm also achieves an accuracy of $\epsilon$ in $\mathcal{O}(T\epsilon^{-1})$ function evaluations, but in practice, for large $T$, the Shubert-Piyavskii method would generally be faster as it better exploits the structure of the function. However, if the time horizon $T$ is relatively small, the computational overhead associated with initializing and maintaining the approximations in the Shubert–Piyavskii method might not be justified, and the grid search algorithm may be competitive or even faster.

\section{Proposed Approach}
In this section, we detail the proposed approach of designing  multi-UAV trajectories that minimize $f_o(\cT)$. We begin with describing the area division and path planning algorithms in Sec. \ref{subsec:roi}. Next, Sec. \ref{subsec:optimization} describes the proposed optimization approach.

\subsection{Area Division, and Path Planning}
\label{subsec:roi}

Given an ROI $\mathcal{A} \subset \mathbb{R}^2$ and the workload ratios $w_i$ (collected in the vector $\mathbf{w}$) of each of the $N$ UAVs such that $\sum_{i=1}^N w_i = 1$, we consider the problem of generating $N$ coverage paths. This requires solving two coupled problems: (1) partitioning $\mathcal{A}$ into $N$ disjoint sub-regions, one for each UAV to cover, such that,
\begin{align}
\mathcal{A} = \bigcup_{i=1}^N \mathcal{A}_i, \quad \mathcal{A}_i \cap \mathcal{A}_j = \emptyset \quad \forall i \neq j
\end{align}
where $\text{Area}(\mathcal{A}_i)/\text{Area}(\mathcal{A}) = w_i$, and (2) generating trajectories $\{\x_i(t)\}_{i=1}^N$ that ensure complete coverage of each sub-region, forming the multi-UAV trajectory $\cT$. Traditional decomposition methods such as trapezoidal and Boustrophedon are not suited for the partitioning step as the resulting partitions do not adhere to a specified workload distribution and are not designed to ensure inter-UAV connectivity as discussed in Sec. \ref{radest}. Instead, we utilize the DARP algorithm proposed in \cite{kapoutsis2017darp}. For the trajectory generation step, we will utilize the STC path planning algorithm \cite{gabriely2001spanning}.

To begin with, both DARP and STC algorithms require the ROI to be discretized into a uniform grid of square cells. Let us denote the discretized ROI as $\cR$ containing cells indexed from 1 to $\abs{\mathcal{R}}$. The side length of each cell is set to twice the sensor footprint $\cF(\x_i(t))$, as required by the STC path planning algorithm. The discretization process utilizes a tunable coverage threshold parameter, $\tau \in (0,1]$ which specifies the percentage  of a cell's area that must be inside $\cR$ for it to be included. That is, a grid cell $j$ (which covers the square region $C_j$) is included in the discretized ROI $\mathcal{R}$ based on the following criteria: 
\begin{align}
j \in \mathcal{R} \quad \text{if} \quad \frac{\text{Area}(C_j \cap \mathcal{A})}{\text{Area}(C_j)} \geq \tau.
\end{align}
For instance, if we set $\tau = 1.0$, only fully covered cells may be included in $\mathcal{R}$, while lower values such as $\tau = 0.1$ allow us to include cells that are at least 10\% inside the ROI. The same thresholding technique applies to cells on the boundaries of the NFZs. To maximize coverage efficiency, we apply a shift and rotation operation to the grid following the optimization method in \cite{apostolidis2022cooperative}. This operation minimizes partially covered cells at ROI edges, optimizing the usable coverage area. 


\subsubsection{Area Division} Given $N$ distinct \emph{seed} cells $p_1$, $\ldots$, $p_N$ (collected in the vector $\p$), the DARP algorithm divides $\mathcal{R}$ into $N$ sub-regions using a modified Voronoi partitioning method. Let $\mathcal{R}_i$ denote the set of indices of the cells included in the $i$-th region. The DARP algorithm is initialized with $\cR_i = \{p_i\}$ and iteratively includes a cell $c$ that is (a) a neighbor of the region $\cR_i$, (b)  not part of any other region, and (c) closest to $p_i$. Each region is updated in a round-robin fashion and the process continues until $|\mathcal{R}_i|/|\mathcal{R}| = w_i$ for all $i \in {1,\ldots,N}$. It is remarked that the DARP algorithm is not guaranteed to terminate for every $\p$. Further, the seed positions critically influence the shapes of $\{\cR_i\}$ and hence the quality of the generated trajectories. We will therefore utilize an outer loop to find the optimal $\p$ that minimizes $f_o(\cT)$.

\subsubsection{Path Planning} We utilize the STC path planning algorithm to generate closed-loop coverage paths within each $\mathcal{R}_i$. Together, the $N$ closed-loop paths form the  multi-UAV trajectory $\cT$. While the launch point of each UAV can be any point along the generated STC path, its position $k_i \in \mathcal{R}_i$ is also optimized to minimize $f_o(\cT)$. The launch points are collected into the vector $\k \in \Rn^N$. 

In summary, a trajectory generated by the DARP and STC algorithms depends only on two $N \times 1$ parameter vectors, namely $\mathbf{p}$ and $\k$, and is denoted by $\cT(\mathbf{p},\k)$.

\subsection{Optimal Trajectories}\label{subsec:optimization}
Finding the optimal trajectory amounts to solving
\begin{align}
    \min_{\p,\k} f_o(\cT(\p,\k)) = r(\cT(\p,\k)) + \lambda e(\cT(\p,\k))
\end{align}
where the connectivity radius $r(\cT(\p,\k))$ is calculated as described in Sec. \ref{radest} and $e(\cT(\k,\p))$ is calculated as described in Sec. \ref{enest}. Observe that there are $\binom{|\cR|}{N}$ possible choices for initializing the DARP algorithm and $\prod_i |\cR_i|$ possible launch points for the STC algorithm. Given the large and discrete search space, a joint optimization approach can be inefficient. Instead, we propose a hierarchical strategy that prunes large parts of the search space based on some heuristics. 

The proposed algorithm proceeds in two nested loops. In the outer loop, we select a seed point $\p$ for DARP initialization and subsequently generate the paths using the STC algorithm. Observer that once the paths are generated, the energy $e(\cT)$ of the trajectory does not depend on the launch points. Hence, the inner loop goes through multiple UAV launch points $\k$ while seeking to find 
\begin{align}
    \k(\p) = \arg\min_{\k} r(\cT(\p,\k))
\end{align}
where $r$ can be calculated as described in \ref{radest}. While generally there are $\prod_i |\cR_i|$ possible launch points for $\k$, we utilize two key heuristics to reduce the search space. Firstly, grid search algorithm is utilized to find $\max_{1\leq i \leq T/\Delta t} r(i \Delta t)$. Secondly, all values of $r(t)$ that are calculated so far are maintained, and at any time if any $r(i\Delta t)$ exceeds the median of recorded values, the current inner iteration is skipped. Both optimization levels are implemented using the tree-structured Parzen estimator (TPE) \cite{watanabe2023tree} which is a Bayesian optimization framework.  Fig. \ref{fig:flow} illustrates this nested optimization process. In summary, the optimization process can be compactly written as 
\begin{align}
    \k(\p) &= \arg\min_{\k} r(\cT(\p,\k)) \\
    \p^\star &= \arg\min_{\p} f_o(\cT(\p,\k(\p)))
\end{align}
and is summarized in Algorithm~\ref{algo:mCPP}.

\begin{algorithm}[t]
\caption{Comm-Energy Aware mCPP}\label{algo:mCPP}
\KwIn{ROI $\mathcal{A}$, $N$, trade-off $\lambda$, UAV parameters}
\KwOut{Optimized multi-UAV trajectory $\mathcal{T}^*$}

Discretize $\mathcal{A}$ into grid $\mathcal{R}$\;
Optimize grid alignment\;
\For(\tcp*[f]{Outer loop}){$\ell = 1$ to $N_{\text{darp}}$}{
  Generate seed cells $\mathbf{p}$ via TPE\;
  Apply DARP($\mathbf{p}$) to partition $\mathcal{R}$ into $\{\mathcal{R}_i\}$\;
  Use STC on each $\mathcal{R}_i$ to form paths\;

  \For(\tcp*[f]{Inner loop}){$j = 1$ to $N_{\text{launch}}$}{
    Generate launch points $\k$ via inner TPE\;
    Compute $r(\mathcal{T}(\mathbf{p},\mathbf{k}))$ and prune if too high\;
    Update inner TPE with $(\k,r(\mathcal{T}(\mathbf{p},\mathbf{k})))$\;
  }
  $\mathbf{k}(\mathbf{p}) \gets \arg\min_{\mathbf{k}}\, r(\mathcal{T}(\mathbf{p},\mathbf{k}))$\;
  Update TPE with $(\mathbf{p}, f_o(\p,\k(\p)))$\;
}
$\p^\star \gets \arg\min_{\p} f_o(\p,\k(\p))$\;
\Return $\cT^\star = \mathcal{T}(\p^\star,\k(\p^\star))$
\end{algorithm}

\begin{rem}
For scenarios with severely constrained communication ranges (typically 3--5 times the sensor footprint width), we develop a formation-flying strategy that ensures continuous connectivity. While formation-based coverage has been explored in \cite{formation}, existing approaches do not account for formation adaptations during turns or explicit communication optimization. 

To handle regular ROIs with tight connectivity limits, define a combined formation footprint $\cF_c = \bigcup_{i=1}^N \cF(\x_i(t))$. We use it to generate a single reference trajectory $\cT_c$ via STC, and keep the individual UAV paths as parallel offsets to the main trajectory, so that, $\x_i(t) = \x_c(t) + d_i\,\mathbf{n}(t)$, where $d_i = (2i-N-1)w/2$ is the lateral offset, $w$ is the sensor footprint width, and $\mathbf{n}(t)$ is the normal to the reference path. The connectivity radius $r(\cT_k)$ is optimized as described in Section~\ref{subsec:optimization}. Although restricted to regular-shaped ROIs without NFZs, this approach guarantees connectivity via a maintained formation, making it particularly effective for large-scale missions dominated by severe communication constraints.
\end{rem}

\begin{figure*}[!ht]
    \centering
    \includegraphics[width=\textwidth]{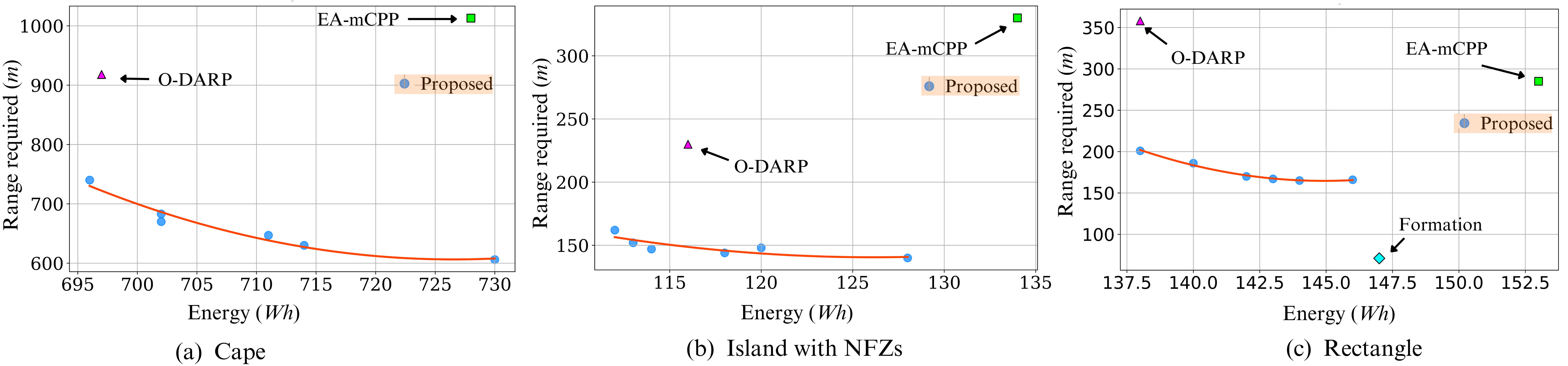}
    \caption{Performance comparison of the proposed algorithm versus state-of-the-art methods across various ROIs}
    \label{fig:perfomance}
\end{figure*}

\section{Evaluation results}
The effectiveness of the proposed mCPP method was evaluated through simulations and real-world tests. The evaluation began with performance validation using the ArduPilot Software-in-the-Loop (SITL) simulator, followed by real-world experiments with three UAVs to assess accuracy and practical performance. 

\subsection{Hardware Specifications and Flight Metrics} \label{sec:flight_matrics}
All simulations were conducted on a computer equipped with a 12th Gen Intel(R) Core(TM) i7-12700K CPU, 32GB RAM, and running Ubuntu 20.04.6 LTS. For the real world experiments, we used a custom-built quadcopter platform, featuring a Raspberry Pi4B as the onboard computer and a Ubiquity Bullet M2 \cite{bulletM} with OpenWRT firmware for wireless mesh networking. We utilized three UAVs operating at an altitude of 45m for both simulation and real-world testing. 
The UAV speeds and the estimated energy consumption parameters are shown in Table \ref{tab:uav_power}. All optimization operations were performed with 3000 DARP trials ($N_{\text{darp}}$) and 1000 shift trials ($N_{\text{launch}}$). With the given computer specifications, these optimizations took approximately 60 minutes to complete.

\begin{table}[htbp]
    \centering
    \renewcommand{\arraystretch}{1.1}
    \caption{UAV Power Consumption by Flight State}
    \label{tab:uav_power}
    \begin{tabular}{p{2.4cm}p{1.6cm}ccc}
        \hline
        \\[-2.5ex]
        \multirow{3}{*}[0.0ex]{UAV State} & \multirow{3}{*}[0ex]{Speed (m/s)} & \multicolumn{3}{c}{Power consumption (W)*} \\[-1ex]
        & & \multicolumn{3}{c}{\makebox[9em]{\hrulefill}} \\[0.1ex]
        & & Mean & Max & Min \\[-.3ex]
        \hline
        \\[-1.50ex]
        Forward flight & 5 & 488 & 510 & 470 \\
        Turning flight & 3 & 509 & 536 & 461 \\
        Hovering & 0 & 492 & 510 & 460 \\
        \\[-2.0ex]
        \hline
    \end{tabular}
    \raggedright
    \\[0.50ex]
    \small{*Measurements taken in normal weather conditions.}
\end{table}


\subsection{Validation in Simulation}
The performance of the proposed method was evaluated through simulations using three UAVs. We compared our method with three existing approaches: Energy-Aware mCPP (EA-mCPP) \cite{energy_aware}, the Optimal DARP (O-DARP) approach \cite{stefanopoulou2024improving}, and a Formation flying strategy \cite{formation}. These comparisons were carried out across multiple ROIs, each varying in complexity and size. The test scenarios included the following ROIs, as detailed in Table \ref{tab:roi_validation}:

\begin{table}[htbp]
    \centering
    \renewcommand{\arraystretch}{1.1} 
    \caption{ROI Specifications for Simulation Validation}
    \label{tab:roi_validation}
    \begin{tabular}{p{2.4cm}p{1.6cm}cc}
        \hline
        \\[-2.2ex]
        \multirow{3}{*}[1.8ex]{ROI Shape} & {Total} & {Sensor} & \multirow{3}{*}[1.8ex]{$\Delta t$ (s)} \\[-.3ex]
        & {Area (m$^2$)} & {Footprint (m$^2$)} & \\[-.3ex]
        \hline
        \\[-1.50ex]
        Cape \cite{shah2022large} & 903,300 & 1225 & 5 \\
        Island & 71,000 & 225 & 1 \\
        Rectangle & 77,000 & 225 & 1 \\
        Simple & 71,000 & 225 & 1 \\
        Complex & 105,000 & 225 & 2 \\
        Island with NFZs & 70,000 & 225 & 1 \\
        \\[-2.0ex]
        \hline
    \end{tabular}
\end{table}

\begin{figure*}[!ht]
    \centering
    \includegraphics[width=\textwidth]{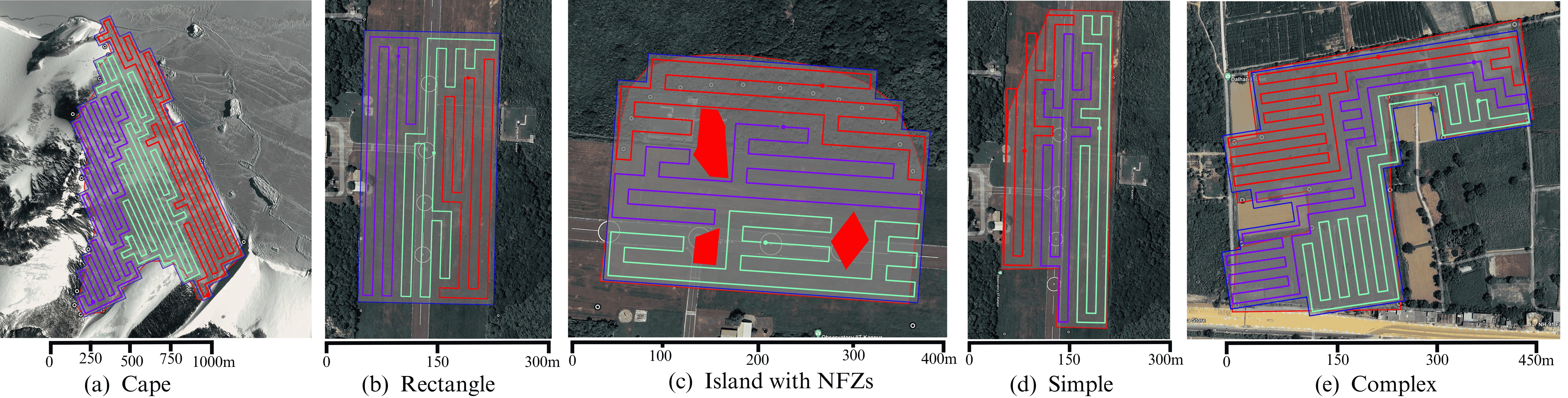}
    \caption{Scenarios used for experimental evaluation of the proposed method. Shade-of-grey polygons represent arbitrary ROI, blue polygon is the processed ROI and solid red polygons are NFZs.}
    \label{fig:scenarios}
\end{figure*}

The visual representation of five of the generated paths is shown in Fig. \ref{fig:scenarios}. The mission altitude used for the Cape shape ROI is 70m above ground lavel and 45 meters for the rest of the ROIs.
Fig. \ref{fig:perfomance} presents the simulation results of the proposed solution across cape, island with NFZs, and rectangular-shaped ROIs, with varying energy-communication trade-off coefficients ($\lambda$). Analyzing the impact of $\lambda$ across different ROIs shows that while the energy-efficient coverage paths have a higher range requirement, the minimum range required paths are not always energy-efficient. In Fig. \ref{fig:perfomance}, the Pareto front is marked in orange, indicating the optimal values tuned for different levels of energy and communication awareness. The energy and communication requirements of the EA-mCPP method are shown with a green rectangle mark, while the requirements of the O-DARP algorithm are indicated with a purple triangle mark. The performance of the formation flying strategy is compared in the rectangular ROI in Fig. \ref{fig:perfomance}(c). Additionally, the connectivity estimation for complex and island ROIs is explained in the supplementary video.

After comparing the performance across different ROIs, the results indicate that our method consistently reduces the communication range requirement to 60\% of what is needed by the existing O-DARP and EA-mCPP methods. Although formation flying is the only existing communication-efficient mCPP technique, it fails to account for energy efficiency. Furthermore, the major limitation of formation flying lies in its dependency on the shape of the ROI and the presence of NFZs. This comprehensive analysis underscores the importance of balancing both energy and communication considerations to achieve optimal performance in multi-UAV coverage path planning.

\subsection{Real World Experiments}
To validate the feasibility of the proposed method in real-world scenarios and assess the accuracy of the simulation, we conducted a multi-UAV cooperative coverage mission with three UAVs. The mission was conducted over the simple polygon ROI mentioned in Table \ref{tab:roi_validation}. 

\begin{figure}[t]
	\centering
	\includegraphics[width=8.5cm]{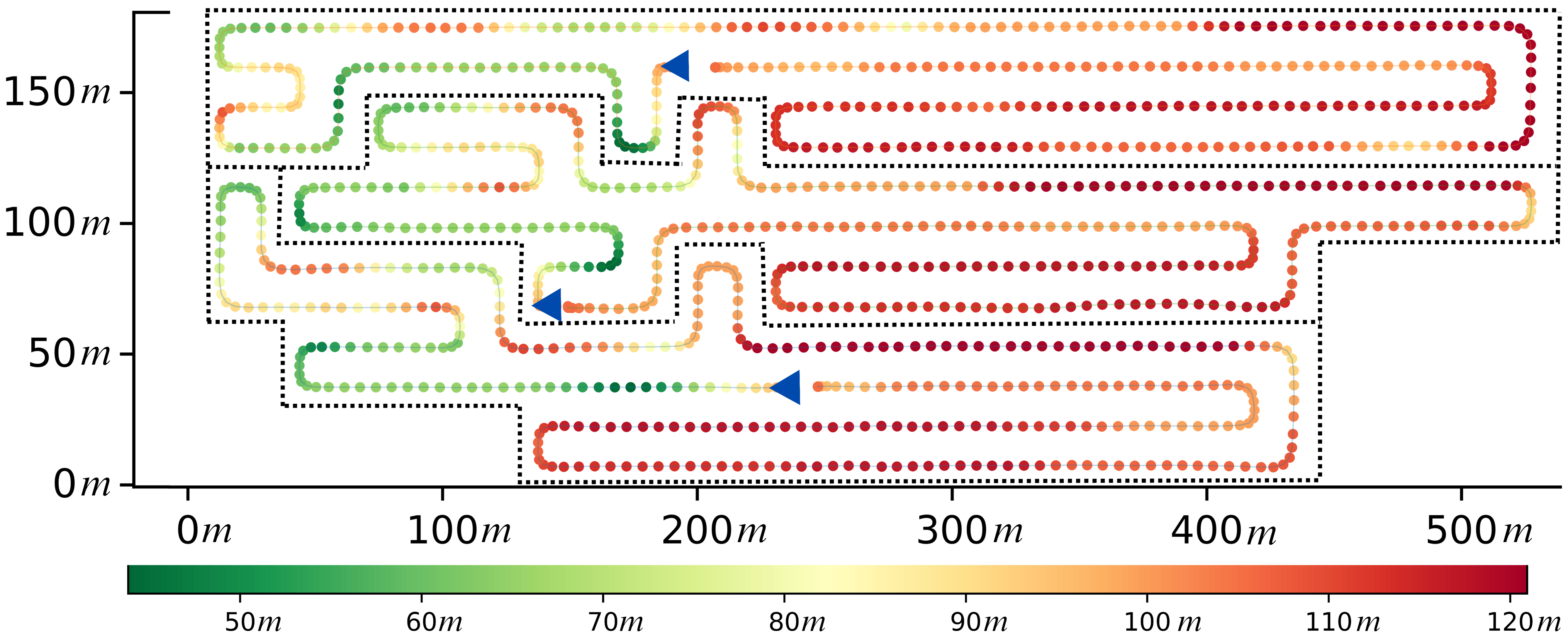}
	\caption{Connectivity strength along UAV paths}
	\label{fig:heatmap}
\end{figure}

The connectivity strength along the actual flight path is shown in Fig. \ref{fig:heatmap}. Blue arrows represent the vehicles' starting positions, with strong connections in green and weak connections in red dots. This visualization helps identify potential connectivity weak spots in the path, allowing for mitigation of connectivity issues, more effective data exchange planning, and enabling adaptive transmission power control for UAVs to conserve energy. To optimize data traffic at these weak spots, adaptive data compression and proactive data caching can be employed, ensuring that critical information is prioritized and transmitted efficiently, minimizing the impact of connectivity issues throughout the flight path.
\begin{figure}[t]
	\centering
	\includegraphics[width=8.5cm]{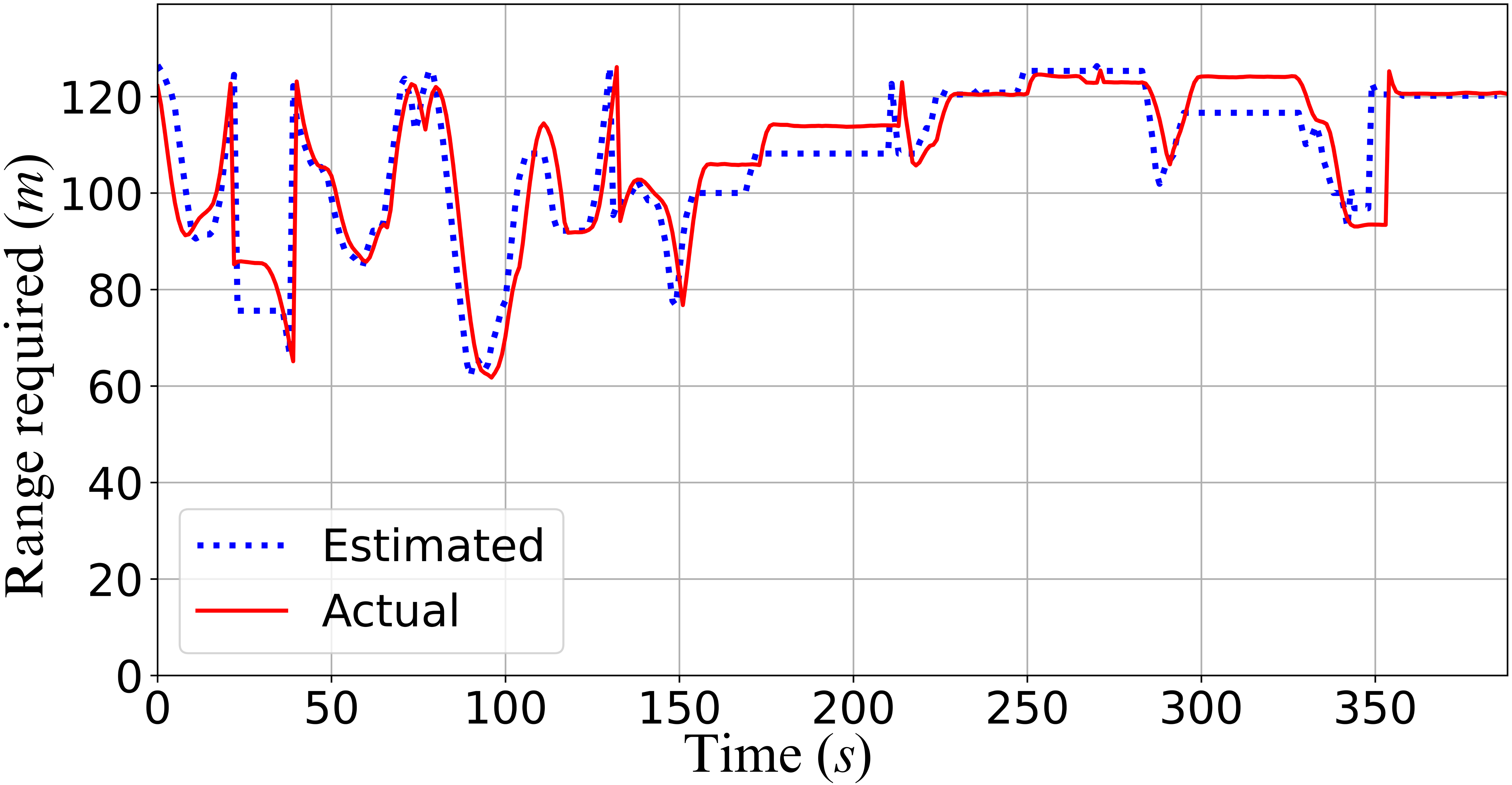}
	\caption{Range requirement vs time}
	\label{fig:reqrd_range}
\end{figure}

The communication range requirements and mission duration from the real-world experiment were compared with the estimated results, as shown in Fig. \ref{fig:reqrd_range}. The estimated minimum range requirement was 125.3 meters, while the real-world experiment showed a slightly lower range requirement of 124.21 meters. This minor difference between the estimated and actual paths is due to the fact that in real-world situations, UAVs take curved turns at corners instead of sharp rectangular turns. Similarly, the estimated mission duration was 385 seconds, but the actual mission took 389 seconds.
This 4-second difference in mission time is attributed to a small error in the turning speed profile of the estimation and actual drones. In the estimation, we did not consider the acceleration or deceleration for switching between forward speed and turning speed. However, actual UAVs have an acceleration model, which introduces a small error of about 150ms during each turn. This error accumulates to 4 seconds over the entire mission.

The real-world experiment demonstrates that the proposed estimation method achieves 99.9\% accuracy in determining the minimum range requirement for the mission (Fig. \ref{fig:reqrd_range}). Minor deviations between the estimated and actual range requirements occur during specific time intervals (155-210s and 300-330s) due to the delay in UAV paths resulting from differences in turning speed, which causes a variation in the relative positions of the UAVs compared to the estimated positions.  This discrepancy is primarily due to approximation techniques and external environmental influences.

\section{CONCLUSION}
We introduce a novel communication and energy-aware multi-UAV coverage path planning method that combines DARP+STC path generation with iterative connectivity assessment optimization. Our approach is the first to generate communication-aware coverage paths for multi-UAV cooperative missions requiring continuous inter-UAV communication. The method incorporates a communication-energy trade-off parameter, allowing users to balance energy and communication awareness during path generation. Simulations demonstrate that our approach outperforms state-of-the-art methods, reducing communication range requirements by 20-60\% and exhibiting better energy awareness compared to existing communication-aware method. The method's feasibility is validated for arbitrary polygonal shapes and polygons with no-fly zones. Real-world experiments with three UAVs confirm the effectiveness of the proposed solution.

\bibliographystyle{IEEEtran}
\bibliography{CAMCPP}


\end{document}